\documentclass[conference]{IEEEtran}
\IEEEoverridecommandlockouts

\usepackage{cite}
\usepackage{amsmath,amssymb,amsfonts}
\usepackage{mathtools}
\usepackage{algorithmic}
\usepackage{graphicx}
\usepackage{textcomp}
\usepackage{xcolor}
\usepackage{booktabs}   
\usepackage{graphicx} 
\usepackage{hyperref}
\usepackage{subfigure}
\def\BibTeX{{\rm B\kern-.05em{\sc i\kern-.025em b}\kern-.08em
    T\kern-.1667em\lower.7ex\hbox{E}\kern-.125emX}}

\newcommand{\mat}[1]{\boldsymbol{#1}}

\newcommand{\norm}[1]{\left\lVert#1\right\rVert}
\begin{document}

\title{Efficient Model-Based Purification Against Adversarial Attacks for LiDAR Segmentation
\thanks{This work has received funding from the EU’s Horizon Europe research and innovation programme in the frame of the CoEvolution project “A comprehensive trustworthy framework for connected machine learning and secured interconnected AI solutions” under the Grant Agreement No 101168560.
}
}

\author{\IEEEauthorblockN{Alexandros Gkillas$^{2}$, Ioulia Kapsali$^{1}$, Nikos Piperigkos$^{2}$, Aris S. Lalos$^{1}$}
\IEEEauthorblockA{$^1$Industrial Systems Institute, Athena Research Center, Patras Science Park, Greece\\
$^2$AviSense.AI, Patras Science Park, Greece\\
Emails: \{piperigkos, gkillas\}@avisense.ai, \{ikapsali\}@isi.gr, lalos@\{athenarc,isi\}.gr 
}}


\maketitle

\begin{abstract}
LiDAR-based segmentation is essential for reliable perception in autonomous vehicles, yet modern segmentation networks are highly susceptible to adversarial attacks that can compromise safety. Most existing defenses are designed for networks operating directly on raw 3D point clouds and rely on large, computationally intensive generative models. However, many state-of-the-art LiDAR segmentation pipelines operate on more efficient 2D range view representations. Despite their widespread adoption, dedicated lightweight adversarial defenses for this domain remain largely unexplored.
We introduce an efficient model-based purification framework tailored for adversarial defense in 2D range-view LiDAR segmentation. We propose a direct attack formulation in the range-view domain and develop an explainable purification network based on a mathematical justified optimization problem, achieving strong adversarial resilience with minimal computational overhead. Our method achieves competitive performance on open benchmarks, consistently outperforming generative and adversarial training baselines. More importantly, real-world deployment on a demo vehicle demonstrates the framework’s ability to deliver accurate operation in practical autonomous driving scenarios.

\end{abstract}


\section{Introduction}

LiDAR is a cornerstone of autonomous vehicle perception, providing precise 3D representations of the environment \cite{li2020lidar}, \cite{mahima2024toward}. A critical application of this modality lies in \textit{semantic segmentation}, wherein each individual point in the point cloud is assigned a categorical label. This step transforms raw data into semantic representations, enabling high-level scene understanding in autonomous systems \cite{jhaldiyal2023semantic}.
Despite significant advances in LiDAR-based perception, recent studies have exposed the vulnerability of deep  networks to adversarial attacks~\cite{xiang2019generating, wen2020geometry}.  These attacks subtly manipulate the structure of point clouds, severely degrading the performance of autonomous perception, initially demonstrated in 3D object detection~\cite{cao2019adversarial, wang2021adversarial}. 
Recently, adversarial attacks have also targeted LiDAR-based segmentation~\cite{zhu2021adversarial}, which is particularly concerning given the central role segmentation plays in decision-making for autonomous vehicles. This raises an urgent need for efficient defense mechanisms specifically tailored to LiDAR segmentation.

To counter adversarial threats in LiDAR-based perception, prior works have explored adversarial training~\cite{lehner20223d, Adv_retraining, Adv_retraining_range} and input purification using generative models such as GANs and diffusion architectures. Initially developed for image synthesis~\cite{ho2020denoising}, generative models have been extended to the 3D domain, where they iteratively reconstruct point clouds through learned denoising dynamics~\cite{PointDP, LiDAR-SPD}. 
Despite their strong denoising capabilities, generative models are typically built with heavy architectures, often far more complex than the downstream segmentation networks they are meant to protect \cite{LiDAR-SPD}. Given that semantic segmentation itself is already a computationally intensive task \cite{jhaldiyal2023semantic}, introducing such heavyweight defense models imposes additional strain on system resources. This added burden significantly limits real-time deployment, particularly on embedded automotive platforms like NVIDIA Orin, where processing power is constrained and latency is critical.
Beyond their computational burden, existing adversarial attacks and defenses have predominantly targeted the raw 3D point cloud domain. However, many state-of-the-art LiDAR semantic segmentation pipelines operate directly on 2D range-view representations, in which the 3D point cloud is projected into structured images \cite{xiao2025revisiting, rethinking, lenet}. This range-view approach is widely adopted due to their computational efficiency and compatibility with real-time inference, making it suitable  for embedded automotive platforms \cite{rethinking}.

In view of this, applying adversarial perturbations in 3D domain and projecting them to 2D often introduces quantization artifacts, weakening the adversarial effect and limiting its impact, as we highlight in Table~\ref{tab:2DVS3D}. To bridge these gaps, we introduce an effective adversarial attack and a lightweight defense framework specifically tailored for LiDAR-based semantic segmentation networks operating in the 2D range-view domain. Our approach first formulates adversarial attacks directly in the 2D range view, injecting structured perturbations that avoid the inefficiencies and artifacts of 3D-to-2D projection. This 2D formulation naturally leads to a novel purification strategy: we pose adversarial mitigation as a constrained optimization problem tailored to the range-view data structure. Leveraging deep unrolling \cite{DeepUnrolling, 10314010}, we transform this optimization problem into an efficient and explainable model-based network, enabling strong robustness.
The main contributions of this paper are:
\begin{itemize}
\item \textbf{2D Range-View Adversarial Attack:} We introduce an adversarial attack specifically designed for LiDAR segmentation networks that operate on 2D range-view data. Conventional 3D attacks lose their effectiveness when the perturbed point clouds are projected onto 2D range images. In contrast, our approach directly injects structured perturbations into the 2D range images, resulting in more effective attacks tailored to the range-view domain.
 
\item \textbf{DU-AP: Lightweight Model-Based Purification:} Building upon our tailored 2D range-view adversarial attack, we introduce the DU-AP (Deep Unrolled Adversarial Purification), a lightweight mitigation module based on deep unrolling principles. By modeling purification as a constrained optimization problem, we design an iterative refinement mechanism that unfolds into an explainable neural network, achieving over $99\%$ fewer parameters than conventional deep learning defenses.
   

\item \textbf{Real-World Deployment on a Vehicle Demo:} The proposed framework is successfully deployed on a demo vehicle equipped with an Nvidia Jetson Orin and a 16-channel LiDAR sensor. Unlike existing SOTA generative purification methods,which are resource-intensive, our model introduces minimal overhead. These findings are further supported by comprehensive evaluations on open benchmarks, demonstrating that our approach efficiently restores segmentation performance with minimal computational burden.

\end{itemize}

\begin{figure}
    \centering
    \includegraphics[scale=0.4]{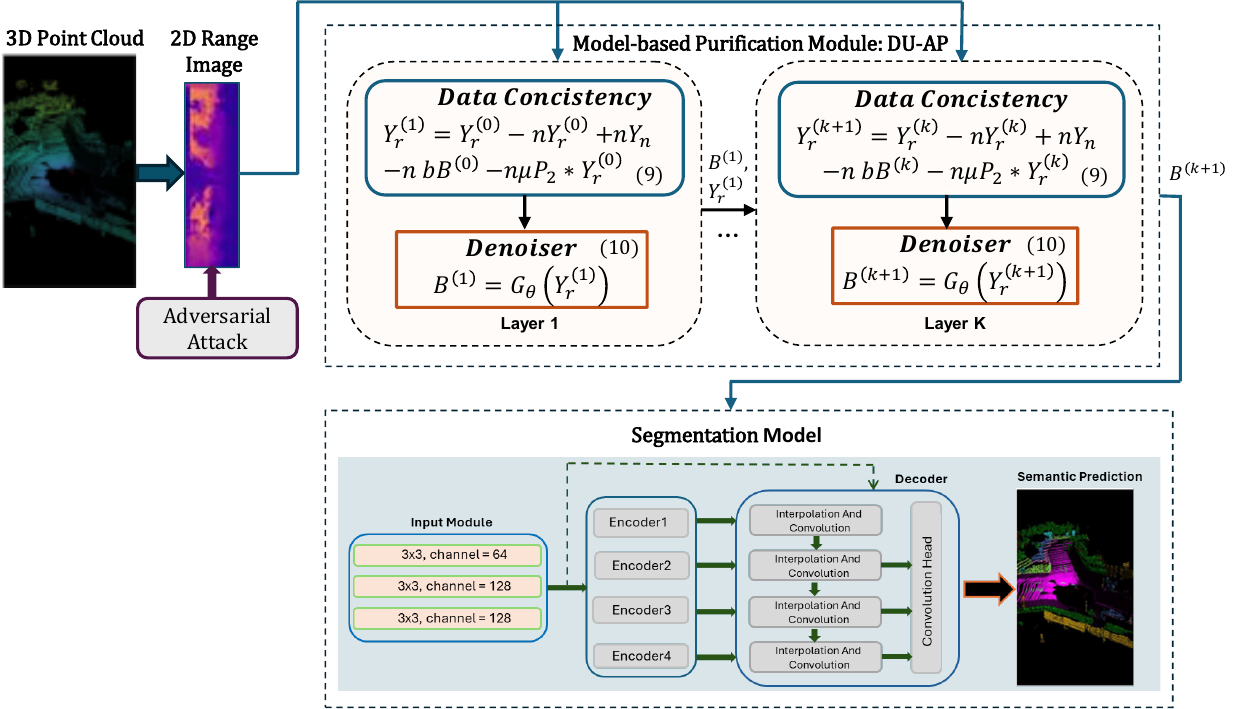}
    \caption{Illustration of the proposed model-based purification framework (DU-AP) for adversarially perturbed LiDAR range images.}
    \label{fig:DU_model}
\end{figure}

\section{Related Work}
\subsection{Adversarial Attacks}
Adversarial attacks on LiDAR perception exploit the vulnerability of deep  networks to subtle, often imperceptible, modifications in the point cloud. Existing attacks can be broadly classified into \textit{point perturbation}, \textit{point injection}, and \textit{point removal} \cite{zhang2024comprehensive}. \textbf{Point perturbation} attacks introduce minimal but strategically optimized displacements to existing points, without altering the global structure of the point cloud \cite{wang2021adversarial}. \textbf{Point injection} techniques, involve the addition of synthetic points into the scene \cite{sun2020towards}. 
Finally, \textbf{point removal} attacks compromise perception by selectively eliminating critical input points identified via saliency analysis \cite{zheng2019pointcloud}. While such attacks have been explored for both 3D pereception systems, such as object detection and segmentation, many recent SOTA segmentation networks process LiDAR data in the 2D range view \cite{rethinking}. 
For these models, adversarial perturbations generated in 3D often lose impact after projection, due to quantization artifacts and geometric distortions~\cite{Adv_retraining_range}. As a result, 3D-domain attacks do not align well with 2D range-view segmentation networks and generally have limited effectiveness (Table~\ref{tab:2DVS3D}). Although 2D range-view architectures are widely adopted, attacks designed specifically for this representation are rare, with only one prior work addressing this challenge~\cite{Adv_retraining_range}. Thus, there is a strong need to develop not only more effective attacks for the 2D range view, but also targeted defense mechanisms that fully exploit the properties of this representation.




\subsection{Adversarial Mitigation}
While adversarial training has been widely adopted as a defense mechanism in LiDAR-based perception, improving the robustness through exposure to crafted attacks \cite{lehner20223d, Adv_retraining, Adv_retraining_range}, it nevertheless suffers from limited generalization to unseen threats and imposes substantial retraining costs \cite{pelekis2025adversarial}.

In contrast, denoising-based purification has emerged as a promising alternative, with generative models standing out for their strong denoising performance. Recent studies have used conditional diffusion models to denoise adversarially perturbed 3D point clouds. These methods operate at the raw point, feature, or voxel level, and aim to enhance the robustness of LiDAR-based 3D object detection~\cite{LiDAR-SPD, LiDARPure}. 
However, these methods are specifically designed to address attacks in the 3D point cloud domain and cannot be directly applied to the 2D range-view representation without \textbf{major} architectural modifications. Moreover, diffusion-based approaches typically require large, complex networks, resulting in computational demands that far exceed those of segmentation models and make deployment on embedded automotive hardware impractical.
This creates a \textit{notable gap} in the literature: although adversarial retraining methods  have been proposed for the 2D range view \cite{Adv_retraining_range}, studies employing purification networks specifically tailored for this domain are lacking. 

One potential direction is the adaptation of generative-based architectures from the RGB image domain: approaches such as APE-GAN~\cite{APE-GAN} and PSGAN~\cite{PSGAN} use generative models to remove adversarial perturbations. More recent frameworks like ATOP~\cite{ATOP} combine adversarial training with generative (GAN-based) purification, while DG-GAN~\cite{DG-GAN} is designed to simultaneously generate and defend against adversarial examples. However, none of these methods are tailored for the unique structure of 2D LiDAR range-view data, nor have they been directly applied in this domain. Furthermore,  GAN-based networks are often unstable during training and computationally demanding at inference, which further limits their practicality for onboard operation in vehicles.
By utilizing the 2D range-view representation of LiDAR data, we are able to express the purification process as a constrained denoising optimization problem. This framework allows us to incorporate regularization terms that capture the unique structural properties of range-view images. We then solve this problem using a deep unrolling strategy, resulting in a model-based network with clear interpretability and strong computational efficiency. As a result, our DU-AP defense can be seamlessly integrated as a lightweight preprocessing module before the segmentation network, providing robust adversarial mitigation.

\section{Proposed method}
Subsection A formulates the attack in the 2D range-view domain. Subsection B introduces the DU-AP, a model-based framework for purifying adversarial inputs.

\subsection{Problem Formulation: 2D Range-view Adversarial Attack}
In LiDAR-based perception, data is acquired as a 3D point cloud $\boldsymbol{P} \in \mathbb{R}^{(N\times3)}$. While conventional adversarial attacks perturb $\boldsymbol{P}$ in 3D, projecting these perturbations to the 2D range view, used by many SOTA segmentation models, introduces quantization artifacts that can weaken their effect. To address this, we design attacks directly in the 2D range view. Let $ \mat{Y_c} \in \mathbb{R}^{C \times M}$ represent the clean range image obtained by projecting $\boldsymbol{P}$, where $C$ is the number of LiDAR channels and $M$ is the horizontal resolution. The adversarial range image is given by:
\begin{equation} 
    \mat{Y_n} = \mat{Y_c} + \mat{\Delta_r}
    \label{eq:adv_problem}
\end{equation}
where $\mat{\Delta_r} \in \mathbb{R}^{C \times M}$ denotes the additive adversarial perturbation. The objective is to degrade the segmentation performance while keeping the perturbation imperceptible, defined as:
\begin{align}
    \min_{\mat{\Delta_r}} \quad & \mathcal{J}(\mat{Y_n}, Z) = -\mathcal{L}_{\text{seg}}(\mat{Y_n}, Z) + \lambda \|\mat{\Delta_r}\|_F^2 \label{eq:attack_obj} \\
    \text{s.t.} \quad & |\mat{\Delta_r}(i, j)| = |\mat{Y_n}(i, j) - \mat{Y_c}(i, j)| \leq \epsilon, \quad \forall (i, j) \label{eq:attack_constraint}
\end{align}
where $\mathcal{L}_{\text{seg}}$ is the segmentation loss, $Z$ is the ground-truth segmentation map, $\lambda$ is a regularization weight, $\mat{\Delta_r}(i, j)$ is the perturbation applied to pixel $(i,j)$ and \( \epsilon \) the maximum allowed deviation per pixel, controlling the perturbation magnitude in terms of $\ell_2$-norm. We solve the  problem adopting Projected Gradient Descent (PGD)~\cite{PGD}, initializing the adversarial input as $ \mat{Y_n}^0 = \mat{Y_c},$ and iteratively updating via:
\begin{equation}
    \mat{Y_n}^{t+1} = \text{Clip}_{\mat{Y_c}, \epsilon}\left( \mat{Y_n}^{t} - \alpha \frac{\nabla_{\mat{Y_n}^t} \mathcal{J}(\mat{Y_n}^{t}, Z)}{\|\nabla_{\mat{Y_n}^t} \mathcal{J}(\mat{Y_n}^{t}, Z)\|_F^2} \right),
\end{equation}
where $\alpha$ is the step size. The clipping operator $\text{Clip}_{\mat{Y_c}, \epsilon}(\cdot)$ enforces the $\ell_2$-norm constraint at each pixel:
\small
\begin{equation}
\text{Clip}_{\mat{Y_c}, \epsilon}(\mat{Y_n}(i, j)) = 
\begin{cases}
\mat{Y_n}(i, j), & \text{if } |\mat{\Delta_r}(i, j)| \leq \epsilon \\
\mat{Y_c}(i, j) + \epsilon \frac{\mat{\Delta_r}(i, j)}{|\mat{\Delta_r}(i, j)|}, & \text{otherwise}
\end{cases}
\end{equation}
\normalsize
This ensures that the perturbation remains bounded. The attack is designed so that the segmentation network $f(\cdot)$ produces incorrect predictions, i.e., $f(\mat{Y_n}) \neq f(\mat{Y_c})$.

\subsection{Model-based Purification: DU-AP}

Following the adversarial degradation defined in Eq.~(\ref{eq:adv_problem}), our objective is to recover a purified range image \( \mat{Y_r} \) from the adversarially perturbed input \( \mat{Y_n} \), such that the segmentation output for \( \mat{Y_r} \) is restored to match that of the clean input \( \mat{Y_c} \). We frame this adversarial purification as a structured denoising problem. Considering the specific structure of the range images derived from 3D LiDAR point clouds,  we propose a novel model-based purification approach, formulated as the following constrained optimization problem:
\begin{equation}
\arg\min_{\mat{Y_r}} \; \frac{1}{2} \| \mat{Y_n} - \mat{Y_r} \|_F^2 + \lambda \mathcal{R}\mat{(Y_r)} + \mu \| \nabla_h \mat{Y_r} \|_F^2 
\end{equation}
Here, the first term is the data consistency and the second term $\mathcal{R}(\cdot)$ is a learnable regularization that promotes properties of the range image. The third term, $\nabla_h \mathbf{Y_r}$, introduces a handcrafted regularizer based on the horizontal gradient, which encourages adjacent pixels—corresponding to neighboring sensor channels—to exhibit similar values.

\subsubsection{Proposed Iterative Solutions}
To solve the problem, the Half Quadratic Splitting \cite{hqs} approach is employed. By using an auxiliary variable \( \mat{B}\), the cost function of HQS is:
\begin{align}
    \mathcal{L}(\mat{Y_r}, \mat{B}) = \frac{1}{2} \norm{\mat{Y_n} -  \mat{Y_r}}_F^2  + \mu\mathcal{R}(\mat{B}) 
   &+ \frac{b}{2}\norm{\mat{B} -\mat{Y_r}}_F^2 \nonumber \\ 
   &+ \mu\norm{\nabla_h \mat{Y_r}}_F^2
   \label{eq:Lagrangian}
\end{align}
where b is a penalty parameter. 
Employing the variable-splitting procedure, we aim to solve these iterative problems:
\begin{subequations}
\begin{align}
    \mat{Y_r}^{(k+1)} = \underset{\mat{Y_r}^{(k)}}{\arg\min}\,&\frac{1}{2} \norm{\mat{Y_n} - \mat{Y_r}^{(k)}}_F^2 
    + \frac{b}{2}\norm{\mat{B}^{(k)} -\mat{Y_r}^{(k)}}_F^2 \nonumber\\ 
    &\quad \quad \quad \quad \quad\quad \quad + \mu\norm{\nabla_h \mat{Y_r}}_F^2,
     \label{eq:updateX}\\
   \mat{B}^{(k+1)} = \underset{\mat{B}}{\arg\min}\, &\lambda \mathcal{R}(\mat{B}^(k)) 
   + \frac{b}{2}\norm{\mat{B}^(k) -\mat{Y_r}^{(k+1)}}_F^2\ . \label{eq:updateZ}
\end{align}
\end{subequations}
\textbf{Data Consistency Module:} The solution of the subproblem (\ref{eq:updateX}) aims to estimate the clear range image using a regularized gradient descent step:
\begin{align}    \label{eq:gdX}
    \mat{Y_r}^{(k+1)}  &= \mat{Y_r}^{(k)} - \eta (-(\mat{Y_n} - \mat{Y_r}^{(k)}) + b\mat{B}^{(k)} \nonumber \\ 
    &+ \mu \nabla_h^T\nabla_h \mat{Y_r}^{(k)})  \\
    &=  \mat{Y_r}^{(k)} - \eta \mat{Y_r}^{(k)} + \eta \mat{Y_n} -\eta b\mat{B}^{(k)}\nonumber
    - \eta  \mu P_2 \star \mat{Y_r}^{(k)}
\end{align}
where \( \eta \) is the learning rate, \( \star \) is the convolution operator and $P_2$ is a learnable 1D circular convolution layer that approximates the horizontal gradient operator $\nabla_h^T\nabla_h$. 

\textbf{Denoising Module:} The solution of the subproblem (\ref{eq:updateZ}) acts as a learnable proximal operator, using a neural network \( \mathcal{G}_\theta \)
to refine the intermediate estimate from the previous step:
\begin{equation}
\mat{B}^{(k+1)} = \mathcal{G}_\theta(\mat{Y_r}^{(k+1)})
\label{eq:denoiser}
\end{equation}
where \( \mathcal{G}_\theta \) is a compact convolutional network designed to remove structured adversarial perturbations from the range image data.  \textbf{Final Iterative Solutions:} The overall purification process alternates between the \textit{data consistency update} in Eq.~(\ref{eq:gdX}) and the \textit{denoising step} in Eq.~(\ref{eq:denoiser}) for a fixed number of iterations.

\subsubsection{Model-based Purification Network - DU-AP}
The goal of the DU-AP is to serve as an effective denoiser, purifying adversarially perturbed range images before segmentation. Since both the denoising and data consistency modules incorporate learnable parameters, the proposed optimization scheme can be restructured as an efficient deep learning architecture. We employ a \textit{Deep Unrolling} approach, where instead of running many iterations of the HQS solver (i.e., Eqs. (\ref{eq:gdX}) and (\ref{eq:denoiser})), we unroll a fixed number of $K$ iterations. This results in a $K$-layer model, called DU-AP where each layer corresponds to a specific step of the proposed optimization, providing  explainability in the network’s function (Fig.~\ref{fig:DU_model}).

\section{Experiments}
To verify the effectiveness and practicality of our approach, we evaluate it on standard benchmark datasets and also conduct real-world experiments using a demo vehicle. 
\subsection{Datasets}
We evaluate our framework on two widely used LiDAR segmentation benchmarks. \textbf{SemanticKITTI}\cite{Semantickitti} is a large-scale dataset collected using a Velodyne HDL-64E sensor in urban driving scenarios. The standard split uses sequences 00–04 for training, 05 for validation, and 06–08 for testing. Each 3D point cloud is projected into a 2D range image of size $64 \times 1024$. \textbf{SemanticPOSS}\cite{Semanticposs} is a segmentation dataset captured with a PANDORA 40-channel sensor in a campus environment. We use sequences 0, 1, 4, and 5 for training, sequence 3 for validation, and sequence 2 for testing. The corresponding 2D range images have a resolution of $40 \times 1800$.
\subsection{Implementation Details}
\textbf{Denoiser architecture}: For the NN $\mathcal{G}_\theta(\cdot)$ in solution (\ref{eq:denoiser}), we used a 5-layer CNN network, where each layer consists of 64 filters with size $3 \times 3$. Moreover, we employed the ReLU as activation and drop-out rate equal to 0.05. \textbf{Proposed DU-AP}: In the proposed model-based DU-AP, the number of unrolling iterations was set to $K=5$, thus leading to 5-layer deep network architecture.
\textbf{Segmentation architecture $f(\cdot)$:}
In our experiments, we employ LENet~\cite{lenet} as the pretrained backbone segmentation network. Designed specifically for range-view LiDAR segmentation, LENet features a compact encoder-decoder architecture with integrated multi-scale attention, balancing computational efficiency and robustness.
\textbf{Evaluation metrics:} Following standard practice in 3D semantic segmentation, we report the Intersection-over-Union (IoU). 
\begin{figure*}
    \centering
    \begin{minipage}{0.25\textwidth}
        \centering
        \includegraphics[width=\linewidth]{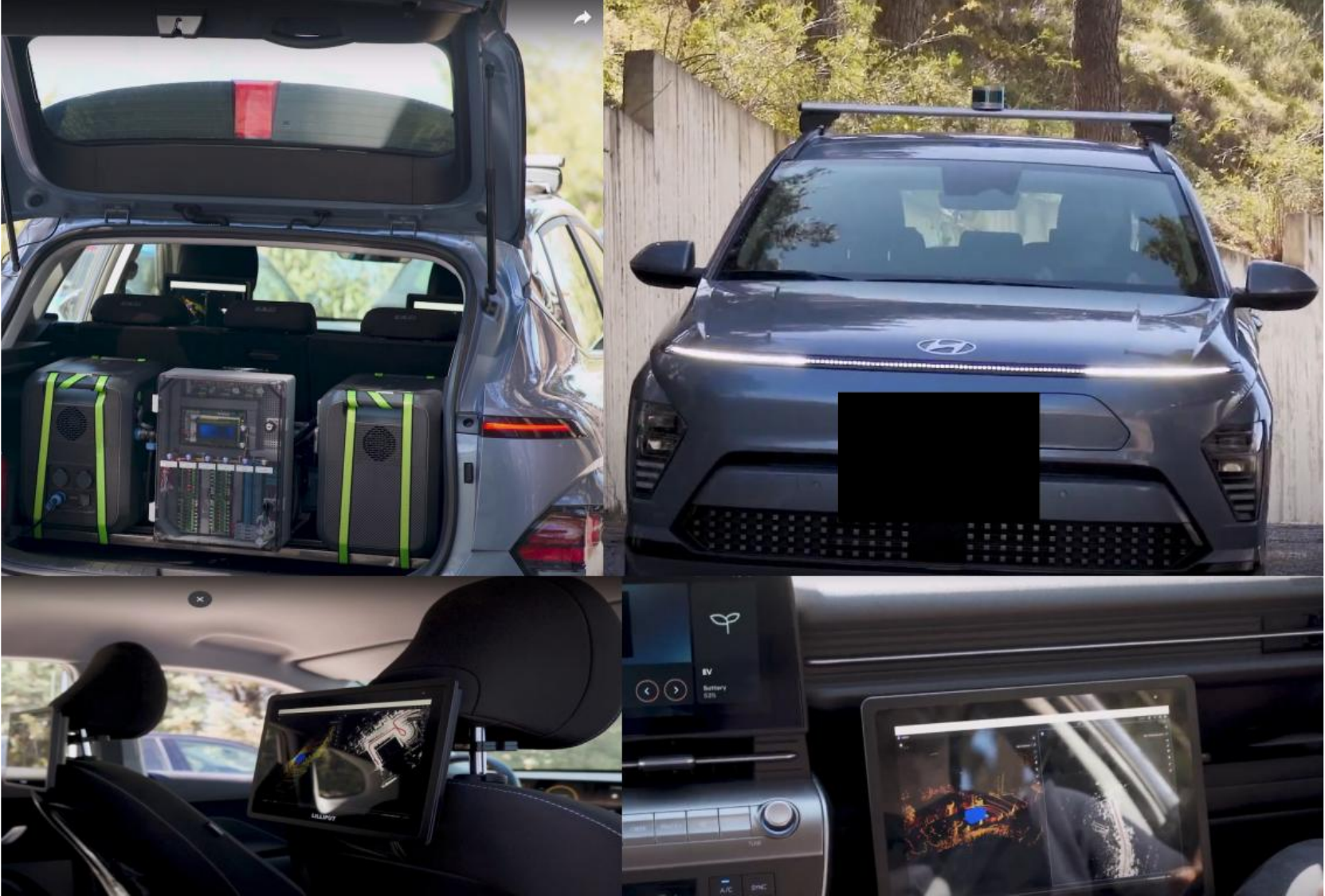}
        \vspace{2pt}
        
        (a)
    \end{minipage}
    \hfill
    \begin{minipage}{0.5\textwidth}
        \centering
        \includegraphics[width=\linewidth]{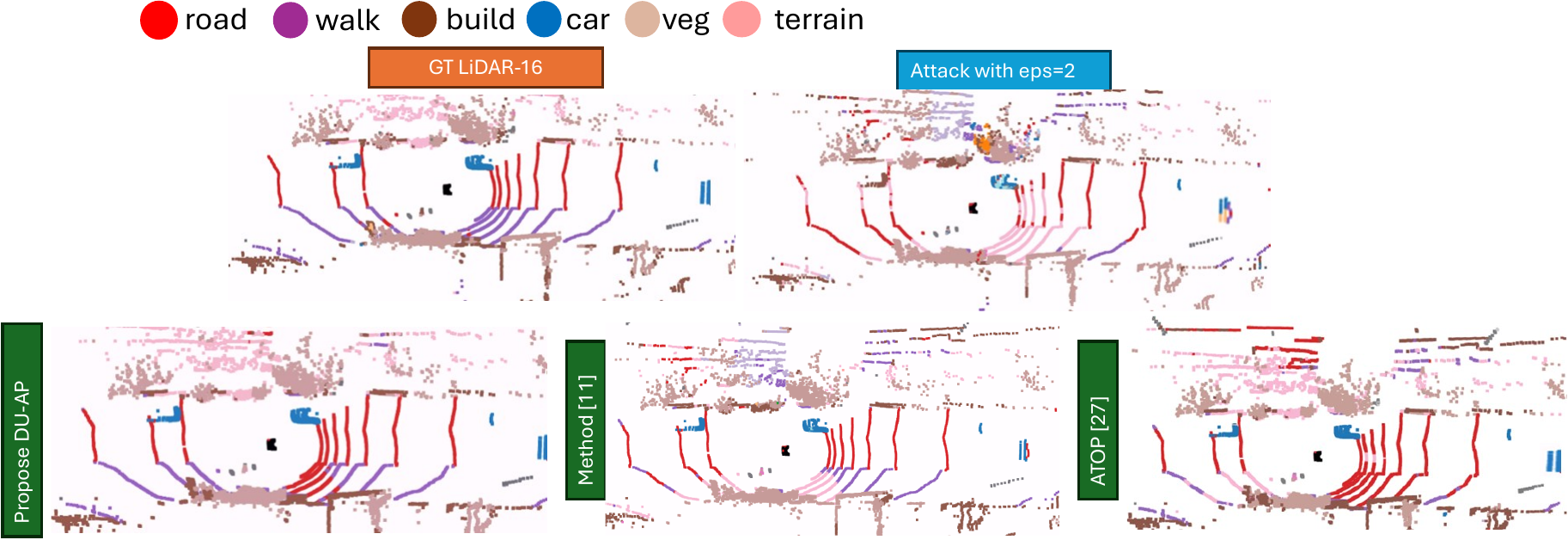}
        \vspace{2pt}
        
        (b)
    \end{minipage}
    \caption{
        (a) Experimental demo vehicle platform. (b) Segmentation results  using a 16-channel Velodyne LiDAR input on the demo vehicle. Four configurations are shown: (i) clean LiDAR-16 input, (ii) adversarial attack with $\epsilon=2$, (iii) the proposed DU-AP, (iv) method [11] and (v) ATOP [27]. Semantic predictions are visualized for various urban scenes. Evaluation uses ground truth labels generated from the segmentation network's predictions on clean input.
    }
    \label{fig:vehicle_and_results}
\end{figure*}
\subsection{Simulation Results: Open datasets}

Table~\ref{tab:2DVS3D} compares the effectiveness of adversarial attacks applied in the 3D versus 2D domains for range-view segmentation, revealing that perturbations applied directly in the 2D range view result in nearly double the performance degradation compared to those crafted in 3D. 
This underscores the superior effectiveness of 2D-domain perturbations, which align directly with the input format of range-view networks and bypass projection-induced artifacts such as quantization and geometric distortion. 
\textbf{Overall IoU Performance:}
Tables~\ref{tab:Comparison_poss} and~\ref{tab:Comparison_Kitti} summarize the average IoU for each method under different perturbation strengths. As expected, adversarial attacks lead to substantial degradation in segmentation performance, with higher $\epsilon$ values causing sharper drops in IoU. For example, in the SemanticPOSS dataset, IoU falls from 0.527 (no attack) to as low as 0.344 under the strongest attack ($\epsilon=9$). The proposed proposed DU-AP  method consistently restores over $60\%$ of the lost IoU across all tested perturbation levels, outperforming adversarial retraining and GAN-based defenses. This resilience is also validated in Table~\ref{tab:per_class_iou}, with notably strong recovery for small classes. In contrast, adversarial retraining provides weaker recovery, and the GAN-based ATOP approach \cite{ATOP}, while competitive, requires substantially more parameters and computational resources.
\textbf{Efficiency and Deployment:}
Importantly, the complexity of GAN-based  method i.e., ATOP \cite{ATOP}, in terms of parameter count, exceeds that of the segmentation network itself. This additional burden makes them prohibitively heavy as a pre-processing step to an already demanding segmentation pipeline, rendering real-time deployment on automotive hardware infeasible. In our case, the proposed DU-AP  module introduces less than $1\%$ additional parameters compared to the segmentation network, enabling real-time performance. 



\begin{table}
\centering
\caption{
Comparison of adversarial attack effectiveness in 2D vs.\ 3D domains for range-view segmentation.
}
\label{tab:2DVS3D}
\small
\resizebox{0.25\textwidth}{!}{%
\begin{tabular}{lccc}
\toprule
 & \textbf{Original} & \textbf{Attack 3D} & \textbf{Attack 2D} \\
\midrule
IoU & 0.527 & 0.444\ (\textbf{--15.9\%}) & 0.365\ (\textbf{--30.9\%}) \\
\bottomrule
\end{tabular}
}
\end{table}

\begin{table}
\centering
\caption{SemanticPOSS: Performance comparison under adversarial attack and defenses. Parameter counts are shown as (defense) + (segmentation) in millions (M).}
\label{tab:Comparison_poss}
\resizebox{0.38\textwidth}{!}{%
\begin{tabular}{lcccc}
\toprule
\textbf{IoU} & $\boldsymbol{\epsilon{=}3}$ & $\boldsymbol{\epsilon{=}6}$ & $\boldsymbol{\epsilon{=}9}$ & 
\begin{tabular}[c]{@{}c@{}}\textbf{Params}\\ \textbf{(M, defense+segm.)}\end{tabular} \\
\midrule
No Attack                                   & 0.527 & 0.527 & 0.527 & -- + 4M \\
Only Attack                                 & 0.398 & 0.365 & 0.344 & -- + 4M \\
Attack \& proposed DU-AP                    & \textbf{0.471} & \textbf{0.460} & \textbf{0.451} & 0.05M + 4M \\
Attack \& Method~\cite{Adv_retraining_range}      & 0.449 & 0.431 & 0.432 & -- + 4M \\
Attack \& ATOP \cite{ATOP}                  & 0.458 & 0.445 & 0.438 & 5M + 4M \\
\bottomrule
\end{tabular}%
}
\end{table}

\begin{table}
\centering
\caption{SemanticKITTI: Performance comparison under adversarial attack and defense.}
\label{tab:Comparison_Kitti}
\resizebox{0.38\textwidth}{!}{%
\begin{tabular}{lcccc}
\toprule
\textbf{IoU} & $\boldsymbol{\epsilon{=}3}$ & $\boldsymbol{\epsilon{=}6}$ & $\boldsymbol{\epsilon{=}9}$ & 
\begin{tabular}[c]{@{}c@{}}\textbf{Parameters}\\ \textbf{(M, defense+segm.)}\end{tabular} \\
\midrule
No Attack                                   & 0.647 & 0.647 & 0.647 & -- + 4M \\
Only Attack                                 & 0.501 & 0.469 & 0.417 & -- + 4M \\
Attack \& proposed DU-AP                    & \textbf{0.609} & \textbf{0.593} & \textbf{0.575} & 0.05M + 4M \\
Attack \& Method~\cite{Adv_retraining_range}      & 0.564 & 0.557 & 0.531 & -- + 4M \\
Attack \& ATOP \cite{ATOP}                  & 0.587 & 0.574 & 0.561 & 5M + 4M \\
\bottomrule
\end{tabular}%
}
\end{table}

\begin{table*}
\centering
\label{tab:per_class_iou}
\small
\caption{\makebox[0.38\textwidth]{Class-wise IoU on SemanticPOSS under Adversarial Attack $\boldsymbol{\epsilon{=}6}$.}}
\label{eps6}
\resizebox{0.8\textwidth}{!}{%
\begin{tabular}{lrrrrrrrrrrrrrr}
\toprule
\textbf{IoU} & \textbf{avg} & \textbf{person} & \textbf{rider} & \textbf{car} & \textbf{trunk} & \textbf{plants} & \textbf{sign} & \textbf{pole} & \textbf{trash} & \textbf{building} & \textbf{stone} & \textbf{fence} & \textbf{bike} & \textbf{ground} \\
\midrule
No Attack         & 0.527 & 0.765 & 0.271 & 0.787 & 0.3 & 0.747 & 0.231 & 0.361 & 0.422 & 0.815 & 0.282 & 0.503 & 0.55 & 0.812 \\
Only Attack       & 0.365 & 0.671 & 0.111 & 0.417 & 0.124 & 0.642 & 0.063 & 0.255 & 0.016 & 0.738 & 0.031 & 0.430 & 0.496 & 0.748 \\
Attack \& proposed DU-AP  & 0.460 & 0.743 & 0.204 & 0.547 & 0.251 & 0.708 & 0.174 & 0.316 & 0.304 & 0.787 & 0.153 & 0.463 & 0.542 & 0.791 \\
Attack \& Method~\cite{Adv_retraining_range} & 0.431 & 0.698 & 0.147 & 0.671 & 0.247 & 0.681 & 0.064 & 0.298 & 0.122 & 0.768 & 0.246 & 0.399 & 0.469 & 0.793 \\
Attack \& ATOP \cite{ATOP} & 0.445 & 0.698 & 0.163 & 0.528 & 0.238 & 0.696 & 0.162 & 0.302 & 0.271 & 0.774 & 0.164 & 0.471 & 0.529 & 0.790 \\
\bottomrule
\end{tabular}%
}
\end{table*}



\subsection{Real-world Deployment on a Demo Vehicle}

To assess practical effectiveness, the framework is deployed on a demo vehicle equipped for autonomous operation.

\subsubsection{Vehicle}
The platform features an NVIDIA Jetson $Orin$ for automotive computing. Environmental sensing is provided by a Velodyne Puck 16-channel LiDAR and an OXTS AV-200 positioning system (GNSS RTK) for centimeter-level localization. The onboard systems are powered by a 4 kW autonomous energy unit (Fig.~\ref{fig:vehicle_and_results}).

\begin{table}
\centering
\caption{Demo vehicle: Performance comparison. Evaluation is based on pseudo ground truth labels derived from the segmentation network's predictions on clean data.}
\label{tab:demo_vehicle_results}
\resizebox{0.34\textwidth}{!}{%
\begin{tabular}{lcccc}
\toprule
\textbf{IoU} & $\boldsymbol{\epsilon{=}1}$ & $\boldsymbol{\epsilon{=}2}$ & $\boldsymbol{\epsilon{=}3}$ &
\begin{tabular}[c]{@{}c@{}}\textbf{Params}\\ \textbf{(M, defense+segm.)}\end{tabular} \\
\midrule
Only Attack                                 & 0.437 & 0.361 & 0.325 & -- + 4M \\
Attack \& proposed DU-AP                    & \textbf{0.541} & \textbf{0.437} & \textbf{0.401} & 0.05M + 4M \\
Attack \& method~\cite{Adv_retraining_range}      & 0.461 & 0.384 & 0.357 & -- + 4M \\
Attack \& ATOP~\cite{ATOP}                  & 0.501 & 0.412 & 0.387 & 5M + 4M \\
\bottomrule
\end{tabular}%
}
\end{table}

\subsubsection{Data and Experimental Setup}
The experiments evaluate the real-time performance of the proposed model-based purification method on a resource-constrained, automotive-grade platform. Data collection was conducted over a 20-minute urban driving route.
We compare segmentation performance across three settings: clean input, adversarial input, and purified input. \textit{Since ground-truth semantic labels are unavailable for the real-world dataset, we use the segmentation model’s predictions on clean data as pseudo ground truth.}

As the segmentation network, we employ a LENet architecture \cite{lenet} pretrained on the SemanticKITTI dataset. Since there are no ground-truth labels available for the real-world dataset, the segmentation network is not fine-tuned on this data, making it more susceptible to adversarial attacks compared to scenarios where the network is trained on in-distribution data. This motivates the use of smaller perturbation strengths ($\epsilon=1,2,3$) in our experiments compared to those used on standard benchmark datasets. Table~\ref{tab:demo_vehicle_results} summarizes the results from the defense mechanisms.  Adversarial retraining is widely used to improve robustness but relies on labeled, in-distribution data for optimal performance. When evaluated out-of-distribution, as in our deployment scenario, its effectiveness drops sharply. This underscores its sensitivity to domain shift scenarios.
In contrast, our DU-AP method leverages an optimization-inspired architecture that inherently improves generalization to new data domains. The proposed approach efficiently recovers segmentation accuracy, as demonstrated by the results in Table~\ref{tab:demo_vehicle_results} and the qualitative examples in Fig.~\ref{fig:vehicle_and_results}. \textbf{Computational efficiency} is crucial for real-world embedded deployment. Generative models e.g., GAN-based ATOP \cite{ATOP} method, perform well on benchmarks but require resource-intensive inference with millions of parameters, making it impractical for real-time operation on NVIDIA Orin. In contrast, our optimization-based approach uses over $99\%$ fewer parameters, confirming its practical deployability in autonomous vehicles.

\section{Conclusions}
In this study, the problem of adversarial robustness in LiDAR-based semantic segmentation was addressed from the perspective of  widely adopted 2D range-view representations. We were motivated by two key limitations in prior work: the quantization artifacts that arise when projecting 3D adversarial perturbations onto 2D, and the high computational cost of existing defense mechanisms. To overcome the first, we proposed a direct attack formulation in the 2D range image space, yielding more effective perturbations. To mitigate such attacks, a novel explainable model-based purification framework was introduced, leveraging the structure of range images to define a constrained denoising optimization problem.  Extensive experimental results on public benchmarks, as well as real-world deployment on an embedded autonomous platform, verified the superiority of the proposed DU-AP model in terms of segmentation accuracy, and computational parameter efficiency.

\bibliographystyle{IEEEtran}
\bibliography{bibl.bib}

\end{document}